\documentclass[Afour,sageh,times]{sagej}

\usepackage{moreverb,url}

\usepackage[colorlinks,bookmarksopen,bookmarksnumbered,citecolor=red,urlcolor=red]{hyperref}
\usepackage{multirow}
\usepackage{forest}
\usepackage{graphicx,subfigure}
\usepackage[normalem]{ulem}
\usepackage{hyperref}

\newcommand\BibTeX{{\rmfamily B\kern-.05em \textsc{i\kern-.025em b}\kern-.08em
T\kern-.1667em\lower.7ex\hbox{E}\kern-.125emX}}

\def\volumeyear{2016}
\begin{document}

\runninghead{Tayyub et al.}

\title{CLAD: A Complex and Long Activities Dataset with Rich Crowdsourced Annotations}

\author{Jawad Tayyub\affilnum{1}, Majd Hawasly\affilnum{2}\thanks{The author was with the University of Leeds at the time of this work}, David C. Hogg\affilnum{1} and Anthony G. Cohn\affilnum{1}}

\affiliation{\affilnum{1}University of Leeds, UK\\
\affilnum{2}Five AI, UK}

\corrauth{Jawad Tayyub, University of Leeds
Woodhouse Ln., Leeds,
LS2 9JT, UK.}

\email{sc12jbmt@leeds.ac.uk}

\begin{abstract}
This paper introduces a novel activity dataset which exhibits real-life and diverse scenarios of complex, temporally-extended human activities and actions. The dataset presents a set of videos of actors performing everyday activities in a natural and unscripted manner. The dataset was recorded using a static Kinect 2 sensor which is commonly used on many robotic platforms. The dataset comprises of RGB-D images, point cloud data, automatically generated skeleton tracks in addition to crowdsourced annotations. Furthermore, we also describe the methodology used to acquire annotations through crowdsourcing. Finally some activity recognition benchmarks are presented using current state-of-the-art techniques. We believe that this dataset is particularly suitable as a testbed for activity recognition research but it can also be applicable for other common tasks in robotics/computer vision research such as object detection and human skeleton tracking.
\end{abstract}


\maketitle

\section{Introduction}
An artificial intelligence system embedded in a human environment, such as a robot, is required to perform \textit{activity recognition} in order to fully comprehend its environment so it could take appropriate steps and make decisions to gain utility. Recognisable/human activities largely vary in complexity, length and expressiveness from short primitive actions lasting for a few seconds, such as `pick', `wave', `approach' etc., to longer activities lasting minutes, hours or days, such as `dinner at a restaurant', `surgery in a hospital', `studying for exams', etc. Currently, most research in human activity recognition has been focused on recognising relatively short activities[]. Though there has been a steady shift towards slightly longer activity recognition, there is still a need for standardised datasets that present longer and more complex activities. In the service robotics domain, this requirement has now become imperative to tackle. A fully autonomous system capable of autonomously running for days or months amongst humans, for example the Strands project~\cite{hawes2016strands}, needs to be equipped with the capability of recognising long-term activities that last hours or days using embedded sensors. However, popular activity datasets, e.g. OPPORTUNITY Activity Dataset~\cite{roggen2010collecting}, offer sensor data from a multitude of inertial and other sensors that are either body worn or installed in a perspective point, such as on the ceiling or corners of rooms, whose outputs are not normally available to an embedded robot in general environments. It is more natural to assume that the robot has to rely primarily on video data from on-board cameras for activity analysis and recognition in these settings. 

In this paper, we propose a dataset called Complex and Long Activities Dataset (CLAD). The dataset can be accessed at this web address : \url{https://doi.org/10.5518/249}. This dataset is i) recorded from the perspective of a mobile robot using a video RGBD camera which is commonly found sensor on a robot and ii) aimed to promote research on long-term complex activities that span longer periods of time than previous datasets. One of the biggest challenges in gathering a datasets consisting of long activity videos comes with the challenge of accurate annotation in a timely and cost effective manner. We therefore further provide a mechanism for obtaining annotations effectively through the use of \textit{crowdsourcing}. Crowdsourced annotations are achieved through a world-wide pool of non-expert users who independently annotate samples of the videos in an objective and unbiased manner. This generates annotations which reflect aspects of true human understanding of the activities within the video, but adds to the complexity from the use of unconstrained natural  language in the annotations which exhibit large variability when describing the videos.

The remainder of this paper is organised as follows: in Section~\ref{relDat} we describe some related datasets that are presently available. Then, in Section~\ref{datDesc} we provide a detailed description of the presented CLAD dataset. In Section~\ref{dsannot} we describe the ground-truth collection process and finally we conclude in Section~\ref{conc}.

\begin{figure*}
\centering     
\subfigure[Raw images from the dataset]{\label{fig:a}\includegraphics[width=80mm]{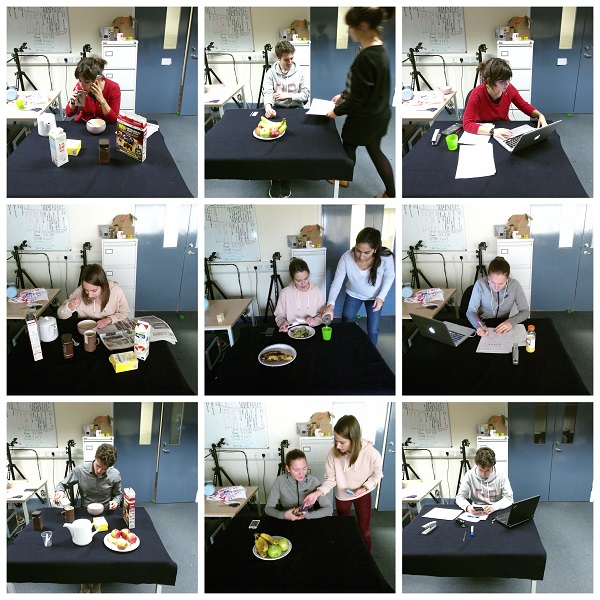}}
\subfigure[Images overlayed with skeleton tracks]{\label{fig:b}\includegraphics[width=80mm]{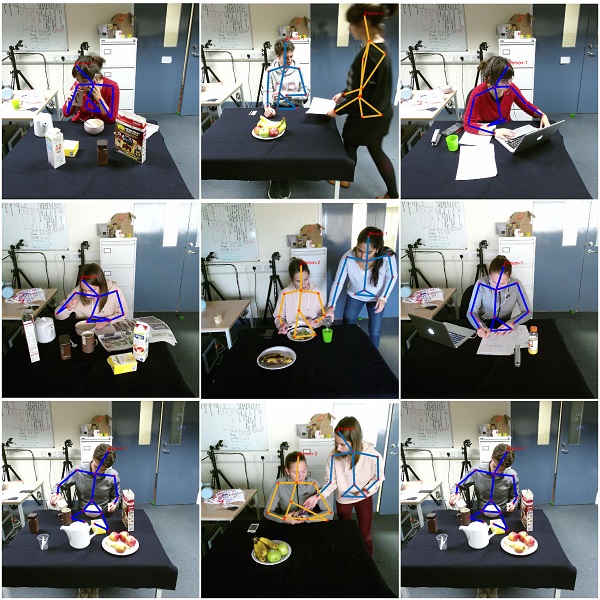}}
\caption{Sample images from the dataset. Each column shows different recordings of a single high-level activity (eating, studying, ...). It can be seen that there is high variability in the performance of these high-level activities between instances.}
\label{sample}
\end{figure*}

\section{Related Datasets}\label{relDat}

Activity recognition has been a popular research direction in robotics. Many activity datasets are openly available for activity recognition research ranging from datasets consisting of simple repetitive action such as jumping, walking etc. (e.g., []) to more complex activities such as cooking food, cleaning microwave etc. (e.g., []). Most available datasets are can be  categorised into three categories: heterogeneous actions, specific actions and others~\cite{chaquet2013survey}. Our dataset falls in the '\textit{activities of daily living}' subcategory of the '\textit{specific actions category}'. Early activity datasets, such as the Weizmann datasets \cite{zelnik2001event,blank2005actions}, captured many instances of various activities which were scripted, recorded in a controlled environment and would rarely be observed naturally. A more natural environment was presented in the KTH dataset~\cite{schuldt2004recognizing} and the Caviar dataset~\endnote{http://homepages.inf.ed.ac.uk/rbf/CAVIAR/caviar.htm}, but the activities presented in these datasets were limited to heterogeneous and repetitive motion such as walking, jumping, running etc., aiming to model the repetitive nature of short activities. UCF datasets (UCF 101, UCF Youtube Dataset)~\cite{soomro2012ucf101,liu2009recognizing} and Hollywood-2 dataset~\cite{marszalek09} were some of the first datasets aimed to model everyday activities in a natural setting such as eating, hugging, kissing etc. from a large number of naturally recorded instances. These datasets were mostly generated from movies or YouTube videos. However, videos in these datasets, though longer than previous datasets, were still short activities where each video only comprised of one basic activity thus temporal segmentation was assumed. \cite{duckworth2017latent} collected a dataset comprising of long videos of human activities recorded by a mobile robot in a kitchen setting. Videos were continuous and comprised of multiple activities. However, the annotations of this dataset were for segmented short actions such as pick up, put down, pour etc. The Cornell Activity Datasets (CAD-120 and CAD-60)~\cite{koppula2013learning,sung2012unstructured} presented challenging and datasets consisting of longer videos than before and with complex activities such as stacking boxes, taking medicine etc. Though this was a benchmarking dataset, it did have some shortcomings. The videos in this dataset only consisted of a single actor and the annotations provided were limited to two levels of granularity: high-level activities per video and sub-level actions within each high-level activity video. In all above mentioned datasets, annotations for activity labels were consistent, extracted from pre-set label lists and collected with a set of well-defined rules in place.  The CLAD dataset presented in this paper is created with particular focus to address some of these shortcomings. 

The dataset presented here consists of much longer activities for example \textit{service at a restaurant}, consisting of multiple actors, contains deep hierarchy of activities and has crowd-sourced freely descriptive annotation i.e. annotators were given the freedom to describe the activities using their own words. Given the advancement towards long-running autonomous robots gathering longer videos of activity data, using crowdsourcing to elicit annotations is becoming popular and therefore such annotations are included in our dataset. To our knowledge, this is the only dataset that presents multiple new challenges to the research community namely, modelling longer naturally occurring and variable activities with many levels of granularity as well as semantic analysis and extraction of information from crowdsourced annotations.     

\section{Dataset Description} \label{datDesc}
The dataset presented in this paper can be accessed from [Dataset URL DOI]. The full size of the dataset available on-line is 222 GB. This is the entire dataset except the point clouds data of the scenes which are available upon request.

\subsection{Recording Setup} 
This dataset comprises of 62 videos recorded at a high resolution of 1920x1080 pixels using a Microsoft Kinect 2 sensor. The sensor was installed at a height of approximately 4.5 feet from the ground. The aim of this view point was to simulate one that a commonly used robot such as a PR2, Tiago, Scitos G5 etc. would normally have when patrolling in an environment. Approximately 8-10 objects relevant to the activity were available to the subjects to use. Five subjects were then employed to act out the scenes. The subjects were told of the top-level activity to act out, after which they were free to use the objects available to act out the task. There were 1-2 subjects involved in each scene. These subjects were carefully selected from outside the faculty having no computer science research experience in order to capture natural acting of activities which is unbiased due to any research knowledge.  Minimal to no instructions were given to the subjects as to how to act out the requested scenes; in fact, they were encouraged to be variable and random in performing the task during different recordings. This allowed for a natural and unbiased data capture of activities exhibiting high variation. In natural setting such as an actual restaurant for a variety of practical and ethical reasons, all the videos were recorded in our research lab; this is a short coming of the dataset which otherwise tries to capture activities in as natural way as possible. Every effort was made to minimise the influence of the environment on the actors by clearing and neutralising any irrelevant items where possible.

\subsection{Dataset Content} 
This dataset comprises of 62 videos of everyday activities ranging from 3 to 10 minutes in length. Only one top-level activity occurs within each of the full videos however numerous sub-level activities occur during the entire video. The top-level activities comprise of three natural scenes of \textit{having breakfast}, \textit{lunch in a restaurant} and \textit{working in an office}. These scenarios were chosen as normally occurring in a \textit{at home}, \textit{restaurant} or \textit{an office} locations likely for a robot to be deployed in. Sample images from the dataset are shown in figure \ref{sample}. Each of the 62 recordings consists of the following data and meta-data:

\begin{itemize}
  \item \textbf{Video:} A low-resolution compressed .mp4 video of the recording. 
  
  \item \textbf{Images:} Uncompressed high definition images of the recording where each image is labelled \textit{Kinect\_\{}frame number\}.\{format\} and has a resolution of 1920x1080 pixels. \{*\} denotes a wildcard for variables in the file names. The images are provided in .png or .jpg formats.
   \item \textbf{Skeletons:} Two state-of-the-art skeleton trackers are used to generate skeleton tracks for all subjects in the recording \cite{wei2016convolutional,rafi2016efficient}. These are provided in the \textit{skeletons} files for each recording. Each skeleton file contains two files corresponding to the two different skeleton trackers used namely \textit{cpm} and \textit{aachen}. Each of the skeleton tracker file consists of files labelled \textit{person\_}\{person's Id\}. Depending on the number of subjects in the video, each person would have a dedicated file of their skeleton tracks identified by the person's Id. In each of the \textit{person\_}\{person Id\} file, there are files labelled \textit{frame\_}\{frame number\}.\textit{txt}. These files are generated per frame and contain a list of (x,y) coordinates that define the location of all joints, of the skeleton onto the image frame in the following order: \textit{head, neck, r-shoulder, r-elbow, r-hand, l-shoulder, l-elbow, l-hand, r-hip, r-knee, r-foot, l-hip, l-knee, l-foot, torso} where \textit{r} and \textit{l} refer to right and left.
  
  \item \textbf{Point Cloud:} This folder consists of the point cloud files labelled \textit{pc\_}\{frame number\}. These are not available on-line but can be acquired upon request. These are not uploaded due to their large file sizes.
  
  \item \textbf{Annotations:} This folder contains annotations of the recording acquired through crowdsourcing. Since there are multiple annotators employed for each video, the annotation files are split by annotator id. The \textit{Annotations} folder therefore contains files labelled \textit{annot}\{annotator's Id\}. Each of the files then contains a list of annotations where each row corresponds to the format: \{subject performing the activity\}, \{activity label/description\}, \{starting frame\}, \{ending frame\}. These files are simple CSV files which are easy and quick to parse and use. More details on how these annotations are gathered is provided in section \ref{dsannot}. Moreover, the \textit{IDs\_of\_Involved\_Subjects.txt} is a file that specifies the unique Ids of the actors performing the activity in that scene. This is just a list of unique numbers corresponding to each actor through out the dataset. 
  
\end{itemize}

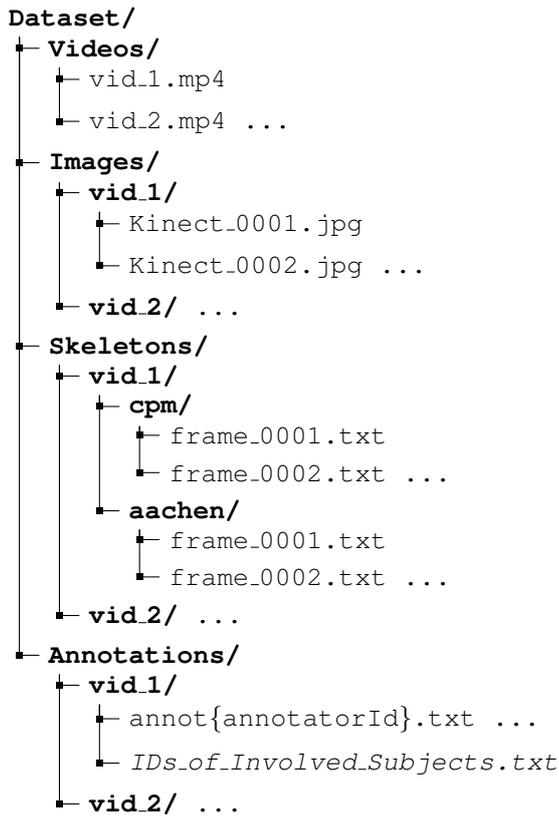
\begin{figure}
\begin{forest}
  for tree={
    s sep=0.02cm,
    font=\ttfamily,
    grow'=0,
    child anchor=west,
    parent anchor=south,
    anchor=west,
    calign=first,
    edge path={
      \noexpand\path [draw, \forestoption{edge}]
      (!u.south west) +(7.5pt,0) |- node[fill,inner sep=1.25pt] {} (.child anchor)\forestoption{edge label};
    },
    before typesetting nodes={
      if n=1
        {insert before={[,phantom]}}
        {}
    },
    fit=band,
    before computing xy={l=15pt},
  }
[\textbf{Dataset/}
  [\textbf{Videos/}
    [vid\_1.mp4]
    [vid\_2.mp4 ...]
  ]
  [\textbf{Images/}
    [\textbf{vid\_1/}
      [Kinect\_0001.jpg]
      [Kinect\_0002.jpg ...]
    ]
    [\textbf{vid\_2/} ...]
  ]
  [\textbf{Skeletons/}
    [\textbf{vid\_1/}
      [\textbf{cpm/}
      	[frame\_0001.txt]
        [frame\_0002.txt ...]
      ]
      [\textbf{aachen/}
      	[frame\_0001.txt]
        [frame\_0002.txt ...]
      ]
    ]
    [\textbf{vid\_2/} ...]
  ]
  [\textbf{Annotations/}
    [\textbf{vid\_1/}
      [annot\{annotatorId\}.txt ...]
      [\textit{IDs\_of\_Involved\_Subjects.txt}]
    ]
    [\textbf{vid\_2/} ...]
  ]
]
\end{forest}
\caption{File structure of the dataset directory}
\label{filestruct}
\end{figure}

The file structure of the entire dataset is shown in figure \ref{filestruct}. 

\section{Dataset Annotations}\label{dsannot}
Annotation of videos involves the identification of activities occurring within the video along with their temporal boundary (their start and end times). Traditionally this task has been performed manually, usually by experts who hold some domain knowledge. The task is expensive, time-consuming and, in case of having domain knowledgeable experts, can result in biased annotations \citep{roggen2010collecting,nguyen2013tagging}. In order to reduce cost and time taken to annotate activities, crowdsourcing platforms are increasingly gaining popularity. A crowdsourcing platform offer a large pool of world-wide workers that are able to perform a human intelligent task (HIT), such as annotation of a video, for a small financial incentive. For large datasets and long videos, as is the case for our dataset, this is a suitable option to attain annotations in a cost-effective and efficient manner. Furthermore, since many annotators are employed to annotate each single video, this helped ensure that a rich and varied perspective of the latent activities were reflected in the annotations. A single person usually tends to annotate videos sequentially one activity after the other in a flat temporal sequence. With multiple annotators, we increased the degree to which we obtained annotations at multiple level of temporal granualirty. Combining these annotations from different workers provides a richer annotation of the video. For our dataset, we make use of a popular crowdsourcing platform called Amazon Mechanical Turk(AMT) \endnote{https://www.mturk.com/mturk/welcome}.

There are however many challenges and design decisions to be taken when developing a system to elicit annotations using crowdsourcing such as deciding on the payment amount, making a clear interface for workers to perform the task, detection and removal of spam or non-diligent workers etc. Overcoming these challenges factors greatly affects the ability to obtain accurate annotations as shown by \cite{nguyen2013tagging}. We will describe our design process next.

\subsection{Interface Design}
For the success of a crowd-sourced annotation system, it is crucial that an easy-to-use interface is provided for workers to submit their annotations through. An image of our interface is shown in figure \ref{interface}. The interface can be divided into the following parts for ease of description:

\begin{figure}
\centerline{\includegraphics[width=0.48\textwidth]{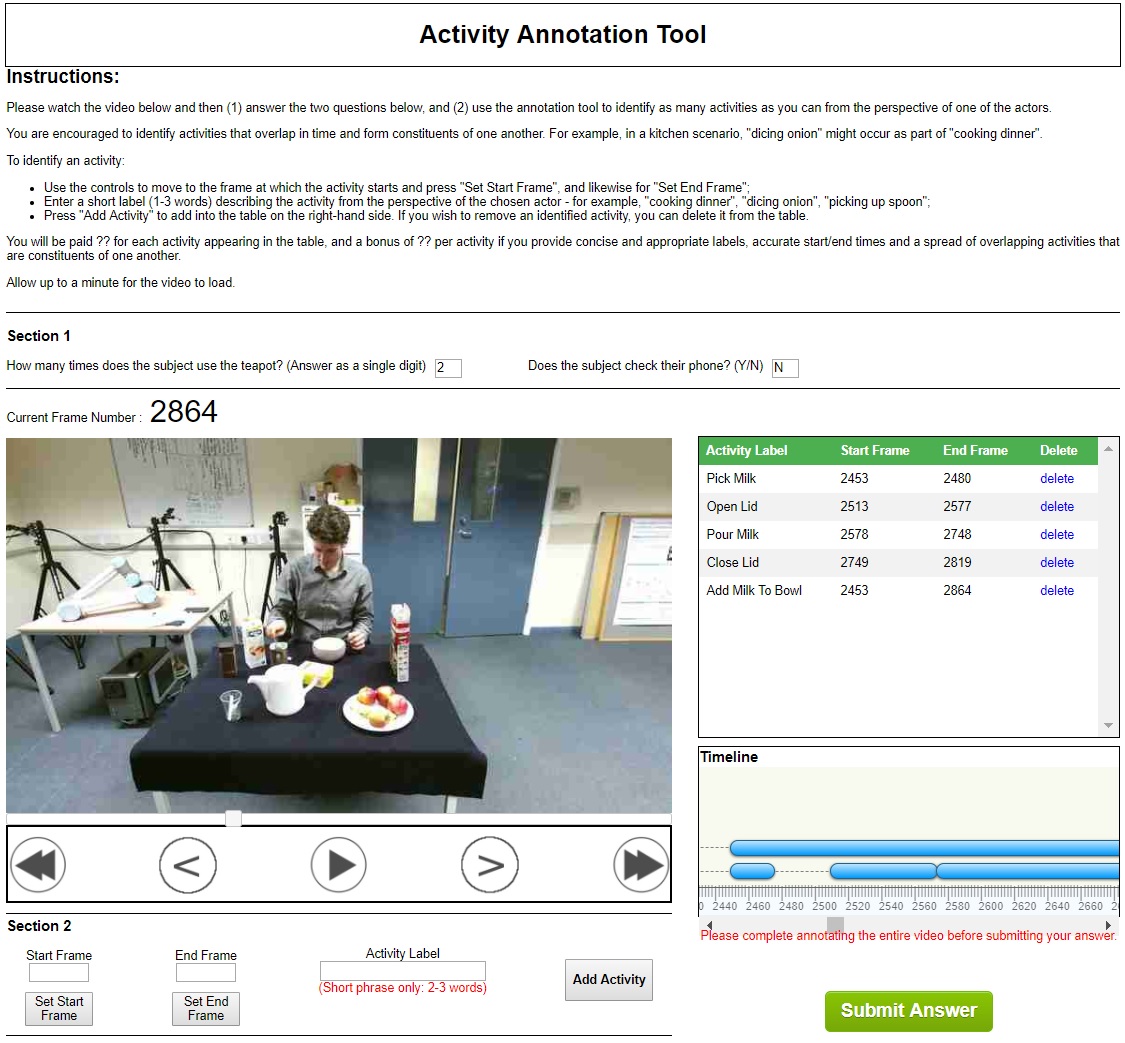}}
\caption{\footnotesize  The web interface developed for Amazon Mechanical Turkers to submit their annotations.} \label{interface}
\end{figure}

\begin{itemize}

\item \textbf{Instructions:} This part simply provides instructions for worker on what is required from them and how to use the interface. The instructions provided in our interface are simply worded since some annotators may not be native English speakers. Also, only the top-level scenario of the activity video is mentioned, no other instructions are given to worker in order to obtain a maximally unbiased set of annotations.

\item \textbf{Video Player:} An easy to use video player with frame skipping control is provided along with the current frame number display. This allows the worker to fine tune the temporal boundaries of identified activities.

\item \textbf{Verification Questions:} An important aspect of the interface is the qualification questions. In order to filter out workers who are inputting spam data or tend to fill in the form with random character inputs, we included two verification questions. These questions are objective, randomly placed and have been pre-answered. A boolean question and a numeric question was used. The aim of the questions was to enforce the user to watch the entire video before answering. For example, a sample boolean question is 'Did the subject use the teapot?' and a sample numeric question is 'How many times did the subject flip the newspaper pages?'. Each numeric question had answers within a range pre-specified? If the worker failed to answer these correctly, they would be disqualified from continuing the task. Through experimenting, we found that inclusion of verification questions helped in discouraging non-diligent user from trying to cheat the task.   

\item \textbf{Annotation Submission Form:} This section allows the user to input the start and end frames of the activity, the activity label and the actor/s performing the activity. The workers are not allowed to type in frame numbers but rather use a button to automatically insert them in the fields. This is done to minimise erroneous input in the frame number's numeric fields. An important design decision made here was to allow the user to freely enter the activity label with minor restrictions such as alphabetical input only and 4 words maximum per activity label to disallow long essay-like descriptions and sentences but produce concise short labels of activities. For example 'consuming coffee' instead of 'the subject is now drinking a cup of coffee'. It is common to preset a list of labels for workers to choose from to avoid ambiguity and variability in natural language but at the same time limits the descriptive power of the worker. We have chosen to allow the workers full control over the choice of words to describe the identified activity to ensure that annotations captured are as expressive without being excessively lengthy. This also removes the requirement of generating a pre-set list of labels for workers to choose from, which can be a time-consuming process in itself.    

\item \textbf{Annotations Table:} In this section the worker can observe the annotations they have identified so far, edit them or delete them. We further impose a restriction of a minimum number of annotations required for each task. Longer tasks would require a higher number of minimum annotations and shorter would require a smaller number. The worker is unaware of this number and is prompted with an error if he/she tries to submit the task without having met the minimum activities identified. Repeated attempts to submit the task with insufficient number of identified activities results in a disqualification warning followed by disqualification.

\item \textbf{Annotation Time-line:} The annotation time-line is simply a graphical representation of the annotations identified along the time-line of the video in frames. This tool is intended to encourage workers to identify parallel and overlapping activities as two overlapping activities will appear on different rows. We are, however, unable to quantify the utility of this feature as there seemed to be no significant change in the quality of annotations received with and without this feature included. Further investigation of the time-line remains as future work.  
\end{itemize}

\subsection{Parameters}

Besides the design decisions that were made regarding the interface used, Amazon Mechanical Turk allows the setting of various crucial parameters that affect the quality of annotations produced. We will briefly describe the important parameters and our choices of their values in this section. An overview of the chosen parameters is shown in table \ref{table:overview}. 

\begin{table}[ht]
\centering
\footnotesize
\begin{tabular}{|p{1.5cm}|c|p{1.5cm}|c|}
\hline
\textbf{Parameter} & \multicolumn{2}{p{3.5cm}|}{Description}  & Value \\ 

\hline
\textbf{Assignment Duration} & \multicolumn{2}{p{3.5cm}|}{Maximum amount of time to finish the task which is 1000 frames long} & 15 minutes\\
\hline
\textbf{Max Assignments} & \multicolumn{2}{p{3.5cm}|}{Number of submissions for the task by unique workers} & 5-7\\
\hline
\textbf{Qualification Requirements} & \multicolumn{1}{p{2cm}|}{Minimum requirement of the worker's qualification} & \multicolumn{1}{p{2cm}|}{Minimum Approved Tasks} & $\geq$20\\
\cline{3-4} & & \% Assignments Approved & $\geq$90\% \\

\hline

\textbf{Reward} & \multicolumn{2}{p{3.5cm}|}{Amount in US dollars paid per task which is 1000 frames long} & up to \$3\\
\hline
\end{tabular}
\caption{Overview of the parameters used for annotation gathering using Amazon Mechanical Turk.  \label{table:overview}
}
\end{table}

It can be seen from the table that there are multiple parameters that require careful thought and experimentation in order optimise the quality of annotations gathered. We will further elaborate some of our choices for the parameters. The 'Assignment Duration' defines maximum time allotted to the worker to finish the task. A pilot study with 6 volunteers was performed to estimate the average maximum needed time (15 minutes) for a task of annotating a 1000 frames video. Given this, any task of $f$ number of frames is give a maximum time of $(f*15)/1000$ minutes following a linear relationship. The 'Reward' amount is computed at a maximum of \$3 for a task of 1000 frames. This amount is split into three parts: i) \$1.5 for passing the verification test and submitting minimum number of required annotations, ii) up to \$1 for quality of labels, start and end times accuracy which were manually checked for a few randomly selected labels and iii) up to \$0.5 for identification of parallel or overlapping activities. Furthermore, bonuses of up to \$1 for a 1000 frame long video were awarded for exceptional work as additional motivation for workers to perform well. Similar to the assignment duration, the reward was linearly adjusted depending on the length of the video in frames. Given the above price model, we ensured that workers were sufficiently motivated to perform the task effectively and within a reasonable amount of time.


\section{Conclusion}  \label{conc}
In this paper, we have presented an activity dataset of naturally occurring daily activities as might be observed by mobile robots. The dataset can be accessed at \url{https://doi.org/10.5518/249}. We further presented the activity annotations gathered through the use of crowdsourcing. We believe this dataset will be useful in robotics and computer vision research. The dataset presents new challenges for long-term autonomous robots systems to comprehend activities they observe. In future, we plan to augment the dataset with object tracks as well as add other activities that involve more subjects interacting and boasts higher complexity. We also plan to record videos in a real-life setting such as a real restaurant or a real office.

\begin{acks}
We acknowledge our colleagues in the School of Computing Robotics
Lab, other schools in the university, and in the STRANDS
project consortium (http://strands-project.eu) for their contributions. We further acknowledge the financial support provided
by EU FP7 project 600623 (STRANDS).
\end{acks}


\theendnotes

\bibliographystyle{SageH}
\bibliography{refs.bib}

\end{document}